# AUTOMATIC TRAINING DATA SYNTHESIS FOR HANDWRITING RECOGNITION USING THE STRUCTURAL CROSSING-OVER TECHNIQUE

Sirisak Visessenee[1,*], Sanparith Marukatat[2], and Rachada Kongkachandra[3]

[1,3] Department of Computer Science, Faculty of Science and Technology, Thammasat University, Klongluang, Patumthani, 12120, THAILAND.
[2] Image Technology Lab., National Electronics and Computer Technology Center, Klongluang, Patumthani, 12120, THAILAND.

## ABSTRACT

*The paper presents a novel technique called "Structural Crossing-Over" to synthesize qualified data for training machine learning-based handwriting recognition. The proposed technique can provide a greater variety of patterns of training data than the existing approaches such as elastic distortion and tangent-based affine transformation. A couple of training characters are chosen, then they are analyzed by their similar and different structures, and finally are crossed over to generate the new characters. The experiments are set to compare the performances of tangent-based affine transformation and the proposed approach in terms of the variety of generated characters and percent of recognition errors. The standard MNIST corpus including 60,000 training characters and 10,000 test characters is employed in the experiments. The proposed technique uses 1,000 characters to synthesize 60,000 characters, and then uses these data to train and test the benchmark handwriting recognition system that exploits Histogram of Gradient: HOG as features and Support Vector Machine: SVM as recognizer. The experimental result yields 8.06% of errors. It significantly outperforms the tangent-based affine transformation and the original MNIST training data, which are 11.74% and 16.55%, respectively.*

## KEYWORDS

*Distortion, Handwriting Recognition, Structural Crossing-Over Technique, Support Vector Machine, Training Data Synthesis*

## 1. INTRODUCTION

Handwriting recognition is a research field that is strongly associated with natural language processing. Its goal is to convert the character images into text that the computer can process. Although various input devices have been developed to provide facilities for users such as keyboard, mouse, digital pen, stylus, and touch screen, however, most people still prefer to write notes on paper with handwriting. To bring them into the computer, users have to waste their time typing them again. Therefore, the development of a handwriting recognition system allows users to easily record and can help bridge the gap between the skilled computer users and those who are not.

In the past, there were a lot of successful researches on handwriting recognition with a high recognition accuracy 90% approximately. Various approaches have been applied in many languages such as English [1-3], Chinese [4,5], Japanese [6], Arabic [7] and Thai [8-10]. The big problem with handwriting recognition research has come from the different handwriting patterns, which are characterized by a specific individual. Also, the handwriting of an individual can vary according to the different emotions and situations. The recognition method used in most research is divided into two main ways.

1. Recognition by the rules (Rule-based Recognition).
2. Recognition by machine learning (Machine Learning-based Recognition).

For the first method, the rules are derived from the experts that are limited, which could not cover all the variety of characters. The second method is more popular because of the following benefits.

- Sophisticated pattern recognition
- Intelligent decisions
- Self-modifying
- Multiple iterations [11]

However, the major problem of the machine learning approach is that the amount of quality training data is insufficient. Garrett Wu [12] said that if we have a lot of data, combined with a simple algorithm, it will be able to overcome a complex algorithm. The good recognizer comes from a good learner. The more experience the learner has, the better the learner is. In this paper, we present a novel technique to automatically synthesize training data for handwriting recognition systems. The paper is structured as follows: Section 2 mentions some existing approaches used to cure the limited training data problem. Section 3 proposes an idea, called the structural crossing-over. Section 4 mentions the experiment settings and their results. The discussion and conclusion are in the last section.

## 2. EXISTING APPROACHES SOLVING INSUFFICIENT TRAINING DATA

There are many researchers attempting to improve the accuracy of handwriting recognition systems by concentrating on training data synthesis. In 2000, M. Mori, et al. [13] proposed a point correspondence technique to generate new samples. They assigned one character as template and then randomly chose a character from the training data set. The point correspondences between the two characters are projected. The generated characters are produced by varying parameters in the defined distortion function, as shown in Fig.1 (a). In 2005, Kambar, Sapargali [14] applied morphing transformation to generate synthetic data for handwritten numeral recognition. He randomly selected two characters from training data set, then three features, i.e. gradient, structural, and concavity, of each character are extracted. Finally, he applied morphing transformation to produce various characters forms, as illustrated in Fig. 1 (b). In 2007, Buyoung Yun, et al. [15] used different tangent vectors to represent different variations. They proposed eight tangent vectors for i.e. scaling, rotation, X-translation, Y-translation, parallel hyperbolic, diagonal hyperbolic, thickness and modified thickness. Figure 1 (c) illustrates the example of applying tangent vectors to the number "2". Although the previous research succeeded in increasing the quantity of training data, they are limited in increasing the variety of patterns for training data.

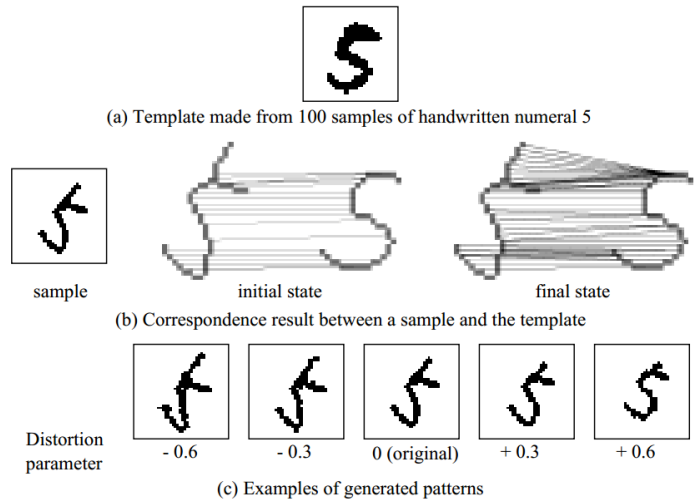

(a) Template made from 100 samples of handwritten numeral 5

sample          initial state          final state

(b) Correspondence result between a sample and the template

Distortion parameter   − 0.6   − 0.3   0 (original)   + 0.3   + 0.6

(c) Examples of generated patterns

(a)

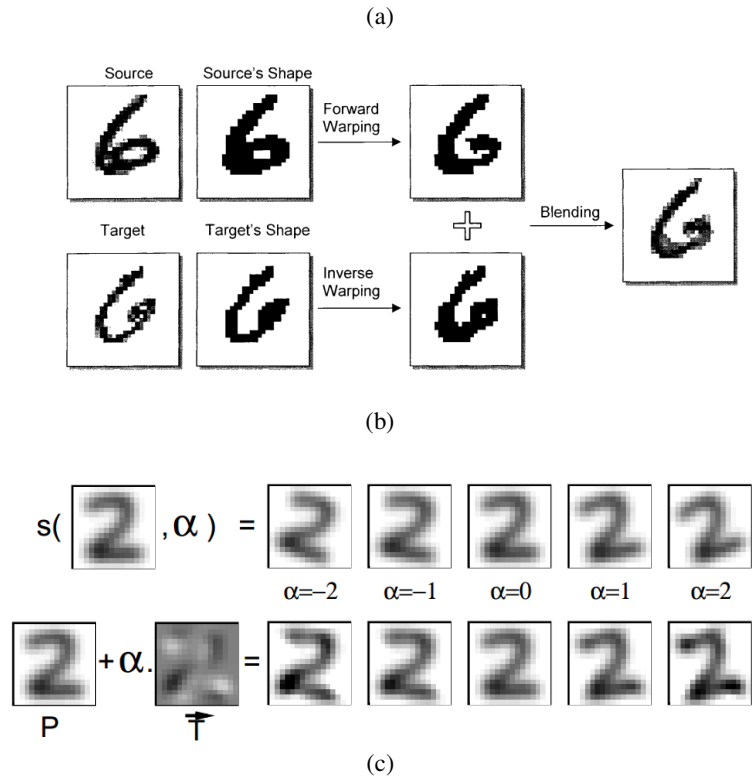

(b)

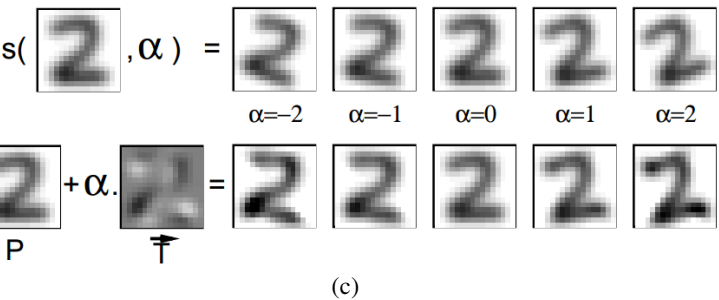

(c)

Figure 1. The generated numeric characters by previous researches

## 3. STRUCTURAL CROSSING-OVER TECHNIQUE

Since the generated characters in the previous works are mostly derived from the source characters by considering the character structure of individual characters. Whether a character is distorted with different angles, its total structure is not much different from the original. In this paper, a couple of characters are selected, some common structures are extracted, the variations between two characters are considered, and finally some new hybrid characters are produced.

The steps in our proposed technique are demonstrated in Fig.2. There are four main steps i.e. preprocessing, crossing-over point finding, structure grouping, and character reproduction.

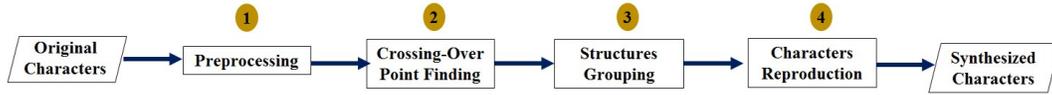

Figure 2. Steps in Structural Crossing-Over Technique

## 3.1 Preprocessing

To generate the new training data, at least two characters are required. They are preprocessed by binarization and thinning, as shown in Fig.3.

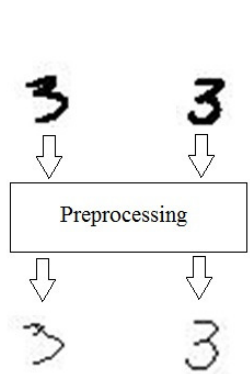

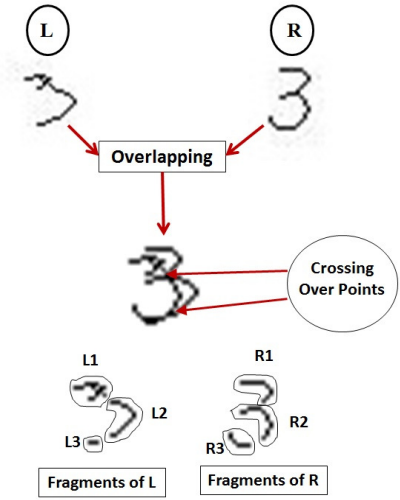

Figure 3. Preprocessing character by "Thinning"     Figure 4. Crossing-Over Point Finding

## 3.2 Crossing-Over Point Finding

This step is to find the common points between two characters. These points are called crossing-over points. Consider in Fig. 4, two original characters, L and R, are overlapped. The coordinate of L is set to (0,0) and the coordinate of R could be varied both in X and Y axis. The number of generated training data depends on the times and the gap size when two characters are overlapped. Each overlapping frame provides one or more intersection points. We then call it "crossing-over point". In Fig.4, there are two points. These points are used as the separator therefore three fragments of each character are obtained.

## 3.3 Structure Grouping

From step 3.2, the different fragments of each character are found. We believe that these various parts can yield various patterns of characters. However, if we make a combination by using these small fragments, some strange characters might occur. To make sure that the synthetic characters are the same as the original character, the structure of the character should be gained. Therefore, these fragments are then overlapped again to find the significant structure of each character. From Fig. 5, six fragments i.e. L1,L2,L3,R1,R2,and R3 are overlapped and then output two main structures. Dilation process is then used to make the structures clear.

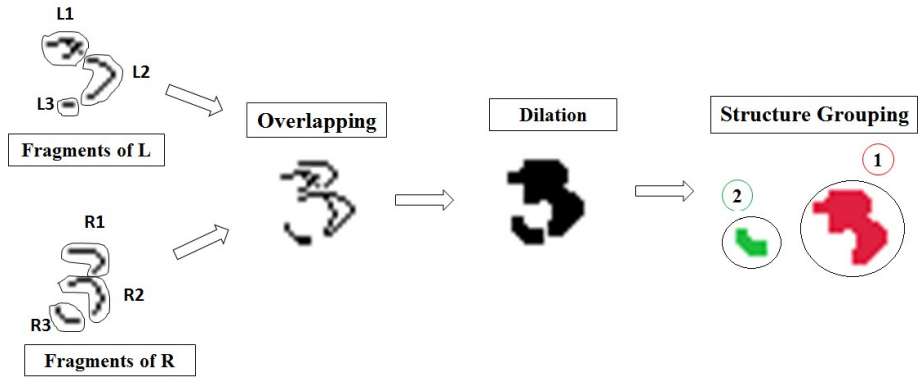

Figure 5. Structure Grouping Process

## 3.4 Characters Reproduction

After the structure fragments of original characters are obtained, the character reproduction step is activated. The new characters are synthesized by using three parts i.e. one fragment from L, crossing-over points, and one fragment of R. The fragment L and R should have different structures. Figure 6 illustrated how new "3" are generated. The left "3" character is synthesized by combining the first structure in L, cross-over points, and the second structure in R while the right "3" uses the second structure in L, cross-over points, and the first structure in R. Finally, the dilation process is employed to enlarge the new characters.

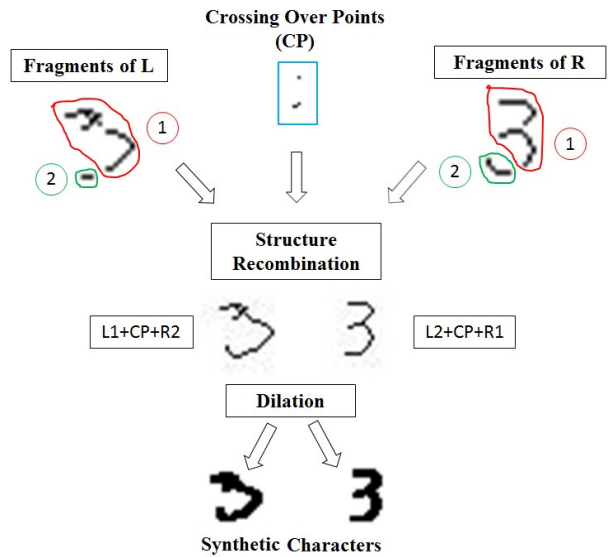

Figure 6. Character Reproduction Process

## 4. EXPERIMENTS AND RESULTS

We hypothesize the proposed "Structural Crossing-over" technique can generate the new characters with a greater variety of patterns than the previous research works. We select the tangent vectors with affine transformation [15] as benchmark. The comparison in terms of character patterns varieties and recognition accuracy are experimented. The standard corpus as MNIST [16] are employed. It includes 60,000 characters for training and 10,000 characters for testing.

Figure 7 shows the appearances of new generated characters after using technique in [15] and our proposed technique. In Fig. 7 (a), ten characters in the right column are generated from the left character. Figure 7 (b) results the new generated characters derived from the proposed technique. The character patterns are more varied.

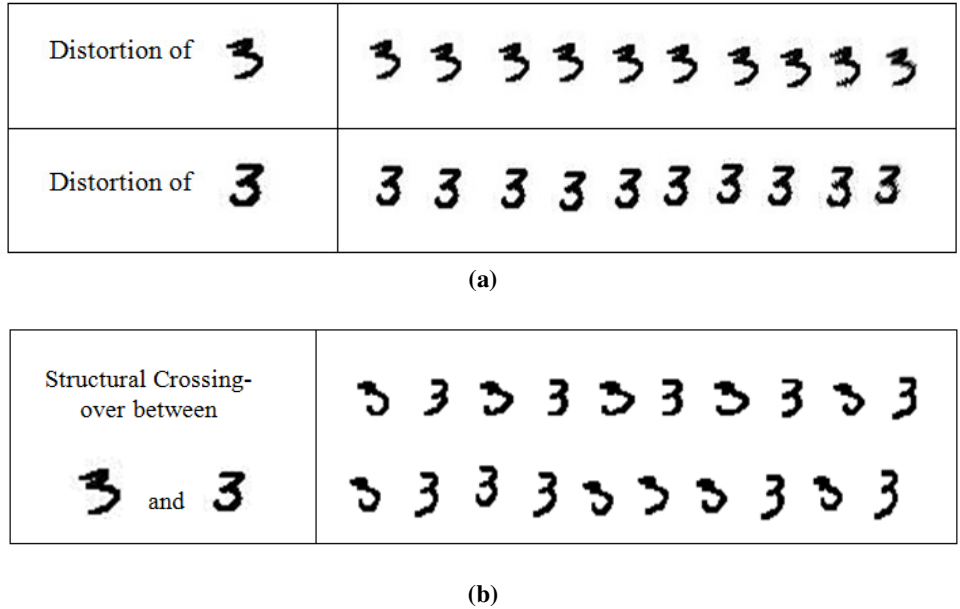

(a)

(b)

Figure 7. Comparative Synthetic Characters between the proposed technique and [15]

With the assumption that if a handwriting recognition is trained by a huge variety of patterns as training data, the recognition accuracy would be increased. We set the experiments by building a benchmark handwriting recognition system that uses the Histogram of Gradient (HOG) as the character features and the Support Vector Machine (SVM) with linear Kernel function and one-against-all approach as multiclass classifier. The benchmark system is trained by 60,000 numeric characters and then is tested with the 10,000 characters from MNIST [15]. The % of recognition error is 16.55.

Instead of using the total 60,000 characters for training, we synthesize the training data from 1,000 numeric characters from MNIST [15]. With the small set of training data, when we applied two synthesis approaches i.e. tangent and structural crossing-over, the % of recognition errors are significantly decreased. The experimental results in the 2nd through the 7th column are the % of recognition errors, which are trained by the synthesized training data vary from 10,000 through 60,000 characters. The proposed technique illustrated the outperformed results.

Table 1 Comparative Percent of Handwriting Recognition Errors proposed technique

| Techniques | Number of training data generated from selected 1,000 from MNIST | | | | | |
|---|---|---|---|---|---|---|
| | 10,000 | 20,000 | 30,000 | 40,000 | 50,000 | 60,000 |
| Tangent | 21.42 | 16.22 | 13.41 | 12.15 | 12.7 | 11.74 |
| Structural Crossing-Over | 10.66 | 9.42 | 9.07 | 8.5 | 8.35 | 8.06 |

# 5. CONCLUSION

The paper presents a novel technique called "Structural Crossing-Over" to synthesize qualified data for training machine learning-based handwriting recognition. The proposed technique can provide a greater variety of patterns of training data than the existing approaches. The tangent vectors with affine transformation [15] are used as a competitive approach. The comparison in terms of character patterns varieties and recognition accuracy are experimented with the MNIST corpus [16]. The Support Vector Machine (SVM) with Histogram of Gradient (HOG) features, linear Kernel function and one-against-all approach for multiclass classification is implemented as a recognition system. The proposed technique can yield the outperformed results both in varieties and accuracy.

## Authors

**Sirisak Visessenee** received a B.Sc degree from Thammasat University, Thailand. Currently, He is Master student in Computer Science in Thammasat University and works as teacher assistant in Computer Science department, Faculty of Science and Technology, Thammasat University. His research interests include Artificial Intelligent, Handwritten Recognition, and Machine Learning.

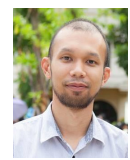

**Sanparith Marukatat** received the PhD degree in computer science from the Paris 6 University (Universite´ Pierre et Marie Curie), France, in 2004. His work concerned handwriting recognition and statistical models for sequence data. He is now with NECTEC, Thailand, where he works on machine learning methods for offline handwriting recognition and for speech recognition.

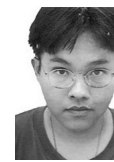

**Rachada Kongakchandra** received the Ph.D. degree in electrical and computer engineering from King Mongkut's University of Technology Thonburi, Thailand. Currently, she works as Assistant Professor at the Computer Science department, Faculty of Science and Technology, Thammasat University. Her research interests include Artificial Intelligent, Natural Language Processing, Semantic Processing, and Machine Learning.

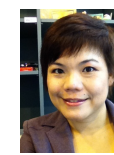